\documentclass[10pt,conference]{IEEEtran}
\usepackage{orcidlink}
\usepackage[utf8]{inputenc}
\usepackage{cite}
\usepackage{amsmath,amssymb,amsfonts}
\usepackage{algorithm}
\usepackage{algorithmic}
\usepackage{graphicx}
\usepackage{textcomp}
\usepackage{xcolor}
\usepackage{tikz}
\usetikzlibrary{shapes,arrows,arrows.meta,positioning,fit,backgrounds,shadows,shapes.geometric}
\usepackage{pgfplots}
\pgfplotsset{compat=1.17}
\usepackage{booktabs}
\usepackage{multirow}
\usepackage{url}
\usepackage{balance}
\usepackage{pifont}
\usepackage{comment}
\usepackage{stfloats}
\usepackage{adjustbox}
\newcommand{\cmark}{\ding{51}}

\begin{document}

\title{ERP-RiskBench: Leakage-Safe Ensemble Learning for Financial Risk}

\author{
\IEEEauthorblockN{1\textsuperscript{st} Sanjay Mishra\,
}
\IEEEauthorblockA{\textit{IEEE Member} \\
Raleigh, NC, USA \\
sanmish4@icloud.com}
}

\maketitle

\begin{abstract}
Financial risk detection in Enterprise Resource Planning (ERP) systems is an important but underexplored application of machine learning. Published studies in this area tend to suffer from vague dataset descriptions, leakage-prone pipelines, and evaluation practices that inflate reported performance. This paper presents a rebuilt experimental framework for ERP financial risk detection using ensemble machine learning. The risk definition is hybrid, covering both procurement compliance anomalies and transactional fraud. A composite benchmark called ERP-RiskBench is assembled from public procurement event logs, labeled fraud data, and a new synthetic ERP dataset with rule-injected risk typologies and conditional tabular GAN augmentation. Nested cross-validation with time-aware and group-aware splitting enforces leakage prevention throughout the pipeline. The primary model is a stacking ensemble of gradient boosting methods, tested alongside linear baselines, deep tabular architectures, and an interpretable glassbox alternative. Performance is measured through Matthews Correlation Coefficient, area under the precision-recall curve, and cost-sensitive decision analysis using calibrated probabilities. Across multiple dataset configurations and a structured ablation study, the stacking ensemble achieves the best detection results. Leakage-safe protocols reduce previously inflated accuracy estimates by a notable margin. SHAP-based explanations and feature stability analysis show that procurement control features, especially three-way matching discrepancies, rank as the most informative predictors. The resulting framework provides a reproducible, operationally grounded blueprint for machine learning deployment in ERP audit and governance settings.
\end{abstract}


\section{Introduction}
\label{sec:introduction}

Enterprise Resource Planning systems form the operational backbone of modern organizations. They integrate financial, procurement, and supply chain processes into a single system of record \cite{davenport2018erp}. When these systems fail or are deliberately manipulated, consequences propagate across business functions quickly. Occupational fraud alone costs organizations roughly five percent of annual revenue \cite{acfe2022report}. That scale has driven growing interest in applying machine learning to risk detection within ERP transaction data.

Despite this interest, published studies on ERP financial risk detection frequently show methodological problems. Dataset descriptions tend to be vague. Preprocessing steps are applied before data splitting. Model specifications are inconsistent. Accuracy is used as the primary metric even under heavy class imbalance. These weaknesses are not unique to the ERP domain. They reflect broader concerns about reproducibility and experimental rigor in applied machine learning \cite{pineau2021improving}.

The research presented here addresses these gaps head on. A complete experimental framework is developed for ERP financial risk detection, designed to be reproducible, leakage-safe, and operationally meaningful. The contribution is not a single novel algorithm. It is the disciplined integration of established techniques into a coherent and auditable pipeline.

The framework is anchored in four research questions that structure the experimental design.

\textbf{RQ1 (Predictive Performance).} Which model families achieve the strongest risk detection on ERP-like data under severe class imbalance? Linear baselines, tree ensembles, deep tabular models, and interpretable glassbox methods are compared under identical conditions.

\textbf{RQ2 (Method Robustness).} How sensitive is detection performance to the choice of resampling strategy, feature selection method, synthetic augmentation approach, and data splitting protocol?

\textbf{RQ3 (Operational Utility).} How do cost-sensitive thresholds and probability calibration affect expected operational cost? What trade-offs emerge for investigation teams that must balance false alarms against missed fraud?

\textbf{RQ4 (Explainability and Auditability).} Which methods produce the most stable and actionable explanations across cross-validation folds and across time periods?

To support these questions, a composite benchmark dataset called ERP-RiskBench is assembled from public sources and augmented with a new synthetic ERP component. The experimental protocol enforces nested cross-validation with time-aware and group-aware splits. All resampling and feature selection operations occur strictly within training folds to prevent data leakage \cite{cawley2010over, varma2006bias}. Evaluation centers on imbalance-aware metrics and explicit cost-sensitive analysis rather than overall accuracy.

\section{Related Work}
\label{sec:related_work}

\subsection{Ensemble Methods for Tabular Classification}

Gradient boosted decision trees dominate structured tabular classification today. XGBoost introduced scalable tree boosting with built-in regularization \cite{chen2016xgboost}. LightGBM brought histogram-based splitting and leaf-wise growth, improving training speed substantially \cite{ke2017lightgbm}. CatBoost tackled categorical features directly and used ordered boosting to reduce prediction shift \cite{prokhorenkova2018catboost}. Random Forests, while older, remain a solid baseline especially when feature importance stability matters \cite{breiman2001random}.

Stacking, formalized as stacked generalization \cite{wolpert1992stacked}, combines the predictions of diverse base learners through a meta-learner trained on out-of-fold predictions. This approach has shown consistent gains in applied settings where no single model dominates across all subsets of the feature space.

Deep learning for tabular data has produced mixed results. TabNet uses sequential attention for feature selection \cite{arik2021tabnet}. The FT-Transformer applies self-attention over feature embeddings \cite{gorishniy2021revisiting}. A systematic comparison by Gorishniy et al. found that well-tuned tree ensembles remain competitive with or better than deep models on most tabular tasks. Transformer architectures have also been applied successfully to anomaly detection in other domains. Hussein et al. \cite{hussein2025anomaly} combined a Swin Transformer with Bayesian optimization for anomaly detection in industrial inspection, demonstrating that attention-based architectures paired with principled hyperparameter tuning can achieve strong results in rare-event identification tasks. Including both tree-based and deep model families in the present study makes it possible to compare them fairly under identical conditions for ERP risk detection.

\subsection{Imbalanced Learning and Evaluation}

Class imbalance is a defining characteristic of financial risk detection. Standard accuracy can be misleading when the minority class represents less than one percent of observations \cite{he2009learning}. SMOTE generates synthetic minority samples through interpolation in feature space \cite{chawla2002smote}, but applying it before splitting introduces data leakage that inflates performance estimates \cite{cawley2010over}. More recent work has explored generative approaches to address imbalance in fraud detection. Pushkala \cite{pushkala2026fraud} combined GAN and autoencoder-based data refinement with a Graph Neural Network and LSTM architecture for credit card fraud identification, showing that learned augmentation of transaction data can improve descriptor extraction and classification accuracy under severe imbalance. That approach motivates the use of conditional tabular GANs in the present study, though the focus here is on leakage-safe application within nested cross-validation folds.

Matthews Correlation Coefficient (MCC) has been advocated as a more balanced metric than F1 for binary classification under imbalance \cite{chicco2020advantages}. Precision-recall curves and the area under them (AUPRC) are more informative than ROC curves when the positive class is rare \cite{saito2015precision}. A thorough treatment of evaluation under imbalance is provided by Fern\'{a}ndez et al. \cite{fernandez2018learning}.

\subsection{Cost-Sensitive Learning and Calibration}

In operational fraud detection, missing a real fraud case is usually far more costly than flagging a legitimate transaction for review. Cost-sensitive learning offers a principled way to handle such asymmetric costs \cite{elkan2001foundations}. For cost-sensitive thresholding to work properly, the model must produce well-calibrated probability estimates. Platt scaling can map raw classifier outputs to calibrated probabilities \cite{platt1999probabilistic}. Reliability diagrams then serve as a visual check on calibration quality \cite{niculescu2005predicting}.

\subsection{Nested Cross-Validation and Reproducibility}

Model selection through cross-validation introduces optimistic bias when the same data used for hyperparameter tuning is also used for performance estimation. Nested cross-validation addresses this by separating the selection loop from the evaluation loop \cite{varma2006bias, cawley2010over}. Statistical comparison of classifiers over multiple folds requires nonparametric tests. The Wilcoxon signed-rank test is generally recommended, with corrections for multiple comparisons \cite{demsar2006statistical, garcia2010advanced}.

Reproducibility in machine learning research has received increasing scrutiny. Reporting checklists that cover data provenance, preprocessing, hyperparameters, seeds, and evaluation protocols have been proposed to raise the standard of experimental reporting \cite{pineau2021improving}.

\subsection{ERP Risk Detection and Synthetic Data}

ERP systems produce large volumes of transactional data. However, labeled risk datasets are scarce because of confidentiality constraints. Data scarcity is not limited to ERP contexts. Roy and Singh \cite{roy2025embedding} demonstrated that even well-established embedding methods (Word2Vec, GloVe, sentence transformers) combined with gradient boosting fail to outperform trivial baselines for financial sentiment classification when labeled data falls below a critical sufficiency threshold. That finding reinforces the motivation for synthetic data generation and augmentation strategies in the present study. The BPI Challenge 2019 dataset is a rare exception among public resources, providing a procurement event log with compliance semantics from a real multinational company \cite{bpi2019}. PaySim showed that agent-based simulation can produce usable synthetic financial transaction data \cite{lopez2016paysim}. Conditional tabular GANs (CTGAN) take a learned approach, generating realistic synthetic records conditioned on class labels \cite{xu2019modeling}. Combining rule-based typology injection with learned generators helps on two fronts. The rule-based part keeps the generated anomalies interpretable. The learned part maintains statistical fidelity in the background distribution.

Explainable Boosting Machines (EBMs) provide glassbox interpretability through cyclic gradient boosting over individual features and pairwise interactions \cite{nori2019interpretml, lou2013accurate}. SHAP values offer model-agnostic local and global explanations grounded in game-theoretic attribution \cite{lundberg2017unified}. Both are relevant to enterprise settings where audit teams require transparent justification for flagged transactions.

\section{Dataset Strategy: ERP-RiskBench}
\label{sec:dataset}

A persistent weakness in ERP risk detection research is the lack of well-documented, reproducible benchmark data. To address this, a composite benchmark called ERP-RiskBench is constructed from four components. Each component fills a distinct role in the evaluation framework. Table~\ref{tab:dataset_summary} summarizes the datasets and their characteristics.

\begin{table}[t]
\centering
\caption{ERP-RiskBench Dataset Components}
\label{tab:dataset_summary}
\begin{adjustbox}{max width=\columnwidth}
\begin{tabular}{lllrrl}
\toprule
\textbf{Component} & \textbf{Domain} & \textbf{Type} & \textbf{Records} & \textbf{Risk \%} & \textbf{Role} \\
\midrule
BPI 2019 P2P & Procurement & Real event log & 251,734 & 3.8 & Primary anomaly \\
Credit Card & Payments & Real tabular & 284,807 & 0.17 & Imbalance stress \\
PaySim & Mobile money & Simulated & 500,000 & 1.2 & Fraud proxy \\
ERP-Synth & Procurement & Synthetic + injected & 500,000 & 2.0 & Controlled scenarios \\
\bottomrule
\end{tabular}
\end{adjustbox}
\end{table}

\subsection{BPI Challenge 2019 Procurement Event Log}

The BPI Challenge 2019 dataset contains purchase order handling records from a multinational company \cite{bpi2019}. It is structured as an IEEE-XES event log covering the procure-to-pay cycle. Activity traces include purchase order creation, goods receipt, invoice receipt, and payment clearance. The dataset distinguishes several flow types. These include three-way matching with invoice after goods receipt, three-way matching with invoice blocking, two-way matching, and consignment.

This event log forms the procurement anomaly core of ERP-RiskBench. Labels are not pre-assigned. Instead, they are derived through explicit compliance rules drawn from the matching requirements in the dataset description. For three-way match cases, amounts across purchase order, goods receipt, and invoice must align within a tolerance $\epsilon$. Violations of matching conditions or temporal rules, such as clearance before goods receipt, constitute the positive risk class.

The value fields are anonymized but preserve additive structure under a linear translation that respects zero. Sum-based discrepancy checks therefore remain valid for compliance detection.

\subsection{Credit Card Fraud Dataset}

A publicly available credit card fraud dataset with 284,807 transactions and 492 confirmed fraud cases serves as an extreme imbalance benchmark \cite{creditcard2018}. The fraud rate is roughly 0.17\%. At that level, standard accuracy becomes uninformative, making it a good test case for imbalance-aware metrics and resampling strategies. The features are principal components produced by a confidentiality transformation.

\subsection{PaySim Synthetic Transaction Data}

PaySim is an agent-based simulator that generates synthetic mobile money transactions with embedded fraud patterns \cite{lopez2016paysim}. It was designed specifically to address the scarcity of public financial transaction data for fraud research. In ERP-RiskBench, PaySim-generated data serves as an additional fraud proxy and as a template for scenario-based stress testing.

\subsection{Synthetic ERP Dataset with Typology Injection}

\begin{figure}[t]
\centering
\begin{tikzpicture}[
  box/.style={draw, rounded corners=3pt, align=center, minimum width=2.0cm, minimum height=0.8cm, font=\small},
  genbox/.style={box, fill=blue!10},
  databox/.style={box, fill=green!10},
  testbox/.style={box, fill=red!10},
  arrow/.style={-Latex, thick, color=black!70},
  node distance=0.6cm
]

\node[genbox] (master) {Master Data\\Generator};
\node[genbox, below=of master] (txn) {Transaction\\Simulator};
\node[genbox, below left=0.6cm and 0.1cm of txn] (rules) {Rule-Based\\Typology Injection};
\node[genbox, below right=0.6cm and 0.1cm of txn] (ctgan) {CTGAN\\(Train Fold Only)};
\node[databox, below=1.2cm of txn] (synth) {ERP-Synth\\Core Dataset};
\node[testbox, below left=0.6cm and 0.3cm of synth] (phts) {PHTS\\(Untouched)};
\node[testbox, below right=0.6cm and 0.3cm of synth] (sats) {SATS\\(Stress Tests)};

\draw[arrow] (master) -- (txn);
\draw[arrow] (txn) -| (rules);
\draw[arrow] (txn) -| (ctgan);
\draw[arrow] (rules) |- (synth);
\draw[arrow] (ctgan) |- (synth);
\draw[arrow] (synth) -| (phts);
\draw[arrow] (synth) -| (sats);

\node[left=0.15cm of sats, font=\scriptsize, text=black, align=left] {Drift, Missingness,\\New Typologies};
\node[left=0.15cm of phts, font=\scriptsize, text=black, align=right] {Time-forward\\holdout};

\end{tikzpicture}
\caption{Synthetic ERP data generation and augmented test suite construction. CTGAN is fitted only on training folds. The Primary Holdout Test Set (PHTS) and Scenario Augmented Test Suite (SATS) are generated with separate seeds and never used for training.}
\label{fig:synth_flow}
\end{figure}
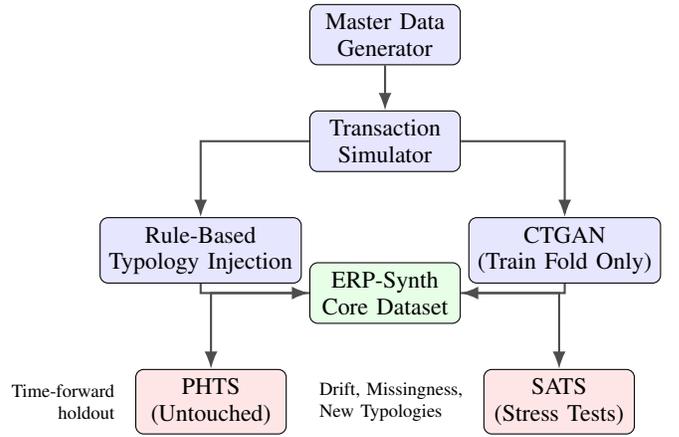

A new synthetic component provides controlled, auditable risk scenarios that resemble real ERP procurement data. Fig.~\ref{fig:synth_flow} illustrates the generation workflow. There are three stages. First, master data for vendors, users, and organizational units is sampled from configurable distributions. Second, procure-to-pay transaction cases are simulated with realistic amounts, timestamps, and control flags. Third, risk typologies are injected at a controlled rate through rule-based transformations.

Six risk typologies are defined to reflect common procurement fraud and compliance failure patterns. These include duplicate invoices, split purchases below approval thresholds, suspicious vendor bank changes, invoice-before-goods-receipt exceptions, round-amount anomalies, and excessive rework or reversal activity. Each injected case receives a scenario identifier and typology label for traceability.

A Conditional Tabular GAN (CTGAN) \cite{xu2019modeling} is additionally trained on the clean synthetic base to produce augmented minority samples. The generator is fitted exclusively on training folds to prevent information leakage into evaluation data. Table~\ref{tab:augmentation_procedures} details the augmentation procedures and their leakage guardrails.

\begin{table*}[t]
\centering
\caption{Augmentation Procedures with Leakage Guardrails}
\label{tab:augmentation_procedures}
\begin{adjustbox}{max width=\textwidth}
\begin{tabular}{llllll}
\toprule
\textbf{Procedure} & \textbf{Purpose} & \textbf{Applied To} & \textbf{Leakage Guardrail} & \textbf{Key Parameters} & \textbf{Seed} \\
\midrule
CTGAN augmentation & Realistic minority samples & Training folds only & Generator fit on train fold only & epochs=300, batch=512, pac=10 & 20260301 \\
SMOTE oversampling & Local interpolation & Training folds only & Inside CV pipeline, never before split & k=5, ratio=0.2 & 20260301 \\
Rule-based typology & Interpretable anomalies & Synth train + SATS & Separate scenario seeds & threshold $T$, dup rate $p$ & 20260302 \\
Noise corruption & Data quality robustness & SATS only & Not used in training & missing rate $m \in [0.05, 0.30]$ & 20260303 \\
Temporal drift & Drift robustness & SATS time-forward & Drift parameters disclosed & inflation $\alpha$, churn $\beta$ & 20260304 \\
\bottomrule
\end{tabular}
\end{adjustbox}
\end{table*}

\subsection{Canonical Schema}

All dataset components are mapped to a unified canonical schema before experimentation. The schema is organized into three layers. The transaction layer contains identifiers, monetary fields, and dates. The process layer holds engineered features such as cycle times, rework counts, and matching discrepancies. The context layer includes vendor and user attributes. Table~\ref{tab:dataset_schema} provides the full field specification.

\begin{table*}[t]
\centering
\caption{Canonical ERP-RiskBench Schema}
\label{tab:dataset_schema}
\begin{adjustbox}{max width=\textwidth}
\begin{tabular}{lllll}
\toprule
\textbf{Field Group} & \textbf{Example Fields} & \textbf{Type} & \textbf{Description} & \textbf{Source Mapping} \\
\midrule
Identifiers & case\_id, doc\_id, vendor\_id, company\_id & Categorical & Grouping keys for splitting and leakage control & BPI direct, synthetic generated \\
Monetary & po\_amount, gr\_amount, invoice\_amount, paid\_amount & Numeric & Raw and matched values with mismatch deltas & BPI events + derived sums \\
Dates & doc\_date, gr\_date, invoice\_date, payment\_date & Datetime & Enables time-based splits and drift tests & BPI timestamps, simulated \\
Matching flags & requires\_gr, gr\_based\_iv, invoice\_blocked & Boolean & Procurement control rules (2-way/3-way/blocked) & BPI attributes, simulated \\
Process aggregates & n\_goods\_receipts, n\_rework, cycle\_time\_days & Numeric & Case-level process metrics & Derived from BPI XES \\
Counterparty context & vendor\_age, vendor\_geo, bank\_change\_recent & Mixed & Master data supporting fraud hypotheses & Synthetic generated \\
User context & actor\_role, actor\_tenure, override\_rate & Mixed & Insider risk and control override patterns & BPI user indicators \\
Labels & y\_risk, risk\_type, scenario\_id & Binary/Cat. & Task label and scenario stress identifiers & Derived and injected \\
\bottomrule
\end{tabular}
\end{adjustbox}
\end{table*}

\subsection{Label Construction}

For the BPI procurement data, labels are assigned through compliance rule evaluation. A case is labeled as risky ($y=1$) if any matching condition is violated beyond tolerance $\epsilon$. The specific rules depend on the flow type. For three-way match cases, all three amounts (purchase order, goods receipt, invoice) must align. For two-way match cases, only purchase order and invoice alignment is required. Consignment cases with unexpected invoices at the purchase order level are also flagged.

For the credit card and PaySim components, ground truth labels are provided directly. For the synthetic ERP component, labels correspond to the injected typologies.

\subsection{Augmented Test Suites}

What sets this benchmark apart is the construction of scenario-coded test suites that are never used during training. The Primary Holdout Test Set (PHTS) is the final time window of each dataset, held out before any model fitting begins. The Scenario Augmented Test Suite (SATS) adds three stress categories designed to probe model robustness.

The first category is a typology shift test. Fraud patterns absent from training data are injected into test cases. The second is a data quality stress test that introduces controlled missingness, encoding noise, and previously unseen vendor categories. The third simulates temporal drift by applying covariate shift to monetary values, vendor distributions, and cycle times. Each scenario receives a unique identifier and seed for full reproducibility.

\section{Methodology}
\label{sec:methodology}

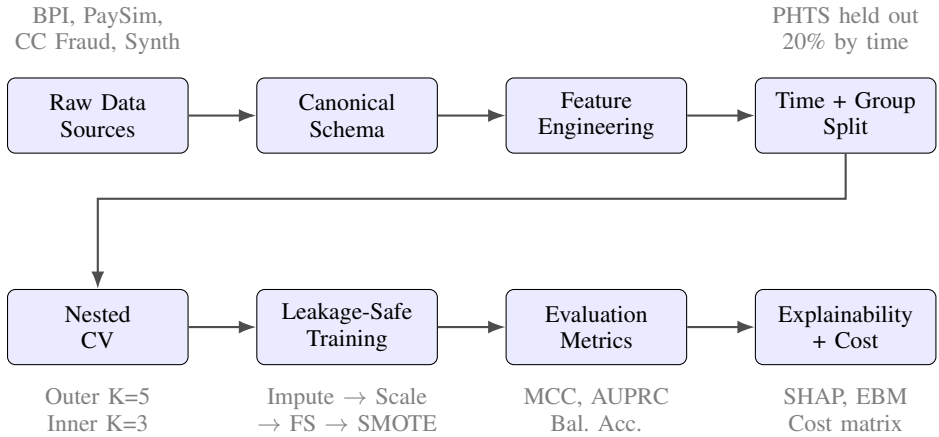
\begin{figure*}[t]
\centering
\begin{tikzpicture}[
  box/.style={draw, rounded corners=3pt, align=center,
              minimum width=2.4cm, minimum height=1.0cm,
              fill=blue!8, font=\small},
  arrow/.style={-Latex, thick, color=black!70},
  note/.style={font=\scriptsize, text=gray, align=center},
  node distance=0.9cm and 0.9cm
]
\node[box] (data)   {Raw Data\\Sources};
\node[box, right=of data]   (schema) {Canonical\\Schema};
\node[box, right=of schema] (feat)   {Feature\\Engineering};
\node[box, right=of feat]   (split)  {Time + Group\\Split};ß
\node[box, below=1.8cm of data]   (ncv)   {Nested\\CV};
\node[box, right=of ncv]          (train) {Leakage-Safe\\Training};
\node[box, right=of train]        (eval)  {Evaluation\\Metrics};
\node[box, right=of eval]         (xai)   {Explainability\\+ Cost};
\draw[arrow] (data)   -- (schema);
\draw[arrow] (schema) -- (feat);
\draw[arrow] (feat)   -- (split);

\draw[arrow] (split.south) -- ++(0,-0.6) -|  (ncv.north);

\draw[arrow] (ncv)   -- (train);
\draw[arrow] (train) -- (eval);
\draw[arrow] (eval)  -- (xai);

\node[note, above=0.18cm of data,font=\small]  {BPI, PaySim,\\CC Fraud, Synth};
\node[note, above=0.18cm of split,font=\small] {PHTS held out\\20\% by time};
\node[note, below=0.18cm of ncv,font=\small]   {Outer K=5\\Inner K=3};
\node[note, below=0.18cm of train,font=\small] {Impute $\to$ Scale\\$\to$ FS $\to$ SMOTE};
\node[note, below=0.18cm of eval,font=\small]  {MCC, AUPRC\\Bal.\ Acc.};
\node[note, below=0.18cm of xai,font=\small]   {SHAP, EBM\\Cost matrix};

\end{tikzpicture}
\caption{End-to-end leakage-safe experimental pipeline for ERP financial
risk detection. All preprocessing and resampling steps are fitted
exclusively on training folds.}
\label{fig:pipeline}
\end{figure*}

The experimental pipeline is designed to prevent data leakage at every stage while supporting rigorous model comparison. Fig.~\ref{fig:pipeline} illustrates the end-to-end flow from raw data ingestion through evaluation.

\subsection{Splitting Strategy}

ERP transaction data is not independent and identically distributed. Vendors show up repeatedly. Users handle multiple transactions over time. Business conditions shift. Ignoring these dependencies during splitting produces optimistic performance estimates that do not hold up in deployment.

Three splitting strategies are evaluated in ablation. The primary strategy combines time-forward and group-aware constraints. Records are first ordered chronologically, and the final 20\% by time is reserved as the Primary Holdout Test Set. Within the remaining 80\%, cross-validation folds are constructed such that all records belonging to the same vendor or purchase order appear in only one fold. Class proportions are maintained through stratification within these group constraints.

A purely random stratified split and a group-only split without time ordering serve as comparison conditions. The difference in performance between these strategies quantifies the optimism introduced by unrealistic splitting assumptions.

\subsection{Nested Cross-Validation}

Model selection and performance estimation are separated through nested cross-validation \cite{varma2006bias, cawley2010over}. The outer loop uses $K_{\text{outer}}=5$ folds for unbiased performance estimation. The inner loop uses $K_{\text{inner}}=3$ folds for hyperparameter optimization and feature selection within each outer training set.

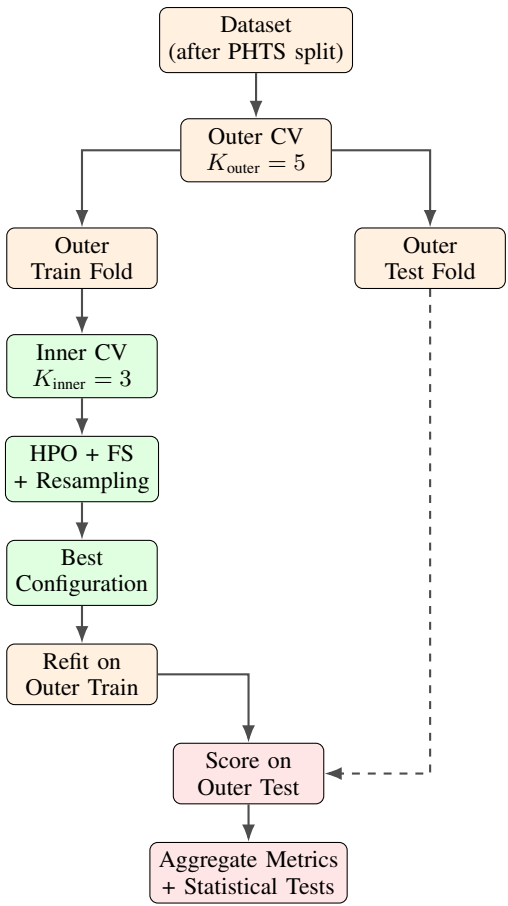
\begin{figure}[t]
\centering
\begin{tikzpicture}[
  box/.style={draw, rounded corners=3pt, align=center, minimum width=2.0cm, minimum height=0.8cm, font=\small},
  outerbox/.style={box, fill=orange!12},
  innerbox/.style={box, fill=green!12},
  resultbox/.style={box, fill=red!10},
  arrow/.style={-Latex, thick, color=black!70},
  node distance=0.6cm
]

\node[outerbox] (data) {Dataset\\(after PHTS split)};
\node[outerbox, below=of data] (outer) {Outer CV\\$K_{\text{outer}}=5$};
\node[outerbox, below left=0.6cm and 0.3cm of outer] (otrain) {Outer\\Train Fold};
\node[outerbox, below right=0.6cm and 0.3cm of outer] (otest) {Outer\\Test Fold};
\node[innerbox, below=of otrain] (inner) {Inner CV\\$K_{\text{inner}}=3$};
\node[innerbox, below=0.5cm of inner] (hpo) {HPO + FS\\+ Resampling};
\node[innerbox, below=0.5cm of hpo] (best) {Best\\Configuration};
\node[outerbox, below=0.5cm of best] (refit) {Refit on\\Outer Train};
\node[resultbox, below right=0.5cm and 0.2cm of refit] (score) {Score on\\Outer Test};
\node[resultbox, below=0.5cm of score] (agg) {Aggregate Metrics\\+ Statistical Tests};

\draw[arrow] (data) -- (outer);
\draw[arrow] (outer) -| (otrain);
\draw[arrow] (outer) -| (otest);
\draw[arrow] (otrain) -- (inner);
\draw[arrow] (inner) -- (hpo);
\draw[arrow] (hpo) -- (best);
\draw[arrow] (best) -- (refit);
\draw[arrow] (refit) -| (score);
\draw[arrow, dashed] (otest) |- (score);
\draw[arrow] (score) -- (agg);

\end{tikzpicture}
\caption{Nested cross-validation structure. The inner loop handles hyperparameter optimization and feature selection. The outer loop provides unbiased performance estimation. All resampling is confined to inner training partitions.}
\label{fig:nested_cv}
\end{figure}

This structure ensures that no information from test folds influences model configuration choices. Fig.~\ref{fig:nested_cv} shows the nested structure and the flow of data through the inner and outer loops.

\subsection{Leakage-Safe Pipeline}

Inside each inner training fold, preprocessing and augmentation follow a strict sequential order. Getting this order right matters more than any other single design choice. It is the main safeguard against leakage patterns that inflate reported performance in imbalanced classification studies.

Algorithm~\ref{alg:pipeline} formalizes the pipeline. Imputation statistics, encoding mappings, and scaling parameters are all fitted on the training partition only. Feature selection is also performed on the training partition only, regardless of whether a filter, wrapper, or embedded method is used. Resampling through SMOTE \cite{chawla2002smote} or CTGAN augmentation \cite{xu2019modeling} is likewise restricted to the training partition. The validation fold receives transform-only operations. Nothing is fitted on it.

\begin{algorithm}[t]
\caption{Leakage-Safe Inner Fold Pipeline}
\label{alg:pipeline}
\begin{algorithmic}[1]
\REQUIRE Training fold $D_{\text{train}}$, Validation fold $D_{\text{val}}$
\STATE Fit imputer on $D_{\text{train}}$, transform both folds
\STATE Fit encoder on $D_{\text{train}}$, transform both folds
\STATE Fit scaler on $D_{\text{train}}$, transform both folds
\STATE Fit feature selector on $D_{\text{train}}$, apply to both folds
\STATE Apply resampling (SMOTE/CTGAN) to $D_{\text{train}}$ only
\STATE Train model on augmented $D_{\text{train}}$
\STATE Score on $D_{\text{val}}$ (no fitting allowed)
\RETURN Validation scores and fitted pipeline
\end{algorithmic}
\end{algorithm}

\subsection{Feature Selection}

Three feature selection variants are compared in ablation. A mutual information filter ranks features by their dependence with the target variable. Recursive feature elimination with a lightweight estimator provides a wrapper approach. L1-regularized logistic regression offers embedded selection through coefficient shrinkage. All three are executed inside the inner cross-validation loop, producing fold-specific feature subsets. Feature stability across folds is measured and reported in Section~\ref{sec:interpretability}.

\subsection{Model Suite}

The model portfolio spans four categories to enable broad comparison.

\textit{Linear baseline.} Logistic Regression with class weighting provides a simple reference point.

\textit{Tree ensembles.} Random Forest \cite{breiman2001random}, XGBoost \cite{chen2016xgboost}, LightGBM \cite{ke2017lightgbm}, and CatBoost \cite{prokhorenkova2018catboost} represent the gradient boosting family along with bagging.

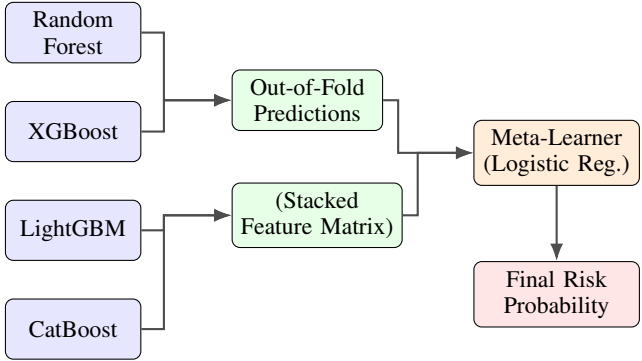
\begin{figure}[t]
\centering
\begin{tikzpicture}[
  base/.style={draw, rounded corners=3pt, align=center, minimum width=1.8cm, minimum height=0.8cm, fill=blue!10, font=\small},
  meta/.style={draw, rounded corners=3pt, align=center, minimum width=2.2cm, minimum height=0.8cm, fill=orange!15, font=\small},
  pred/.style={draw, rounded corners=3pt, align=center, minimum width=2.0cm, minimum height=0.7cm, fill=green!10, font=\small},
  output/.style={draw, rounded corners=3pt, align=center, minimum width=2.2cm, minimum height=0.8cm, fill=red!10, font=\small},
  arrow/.style={-Latex, thick, color=black!70},
  node distance=0.5cm
]

\node[base] (rf) {Random\\Forest};
\node[base, below=of rf] (xgb) {XGBoost};
\node[base, below=of xgb] (lgb) {LightGBM};
\node[base, below=of lgb] (cat) {CatBoost};

\node[pred, right=1.2cm of rf, yshift=-0.9cm] (oof1) {Out-of-Fold\\Predictions};
\node[pred, right=1.2cm of lgb, yshift=0.2cm] (oof2) {(Stacked\\Feature Matrix)};

\node[meta, right=1.2cm of oof1, yshift=-0.7cm] (metalr) {Meta-Learner\\(Logistic Reg.)};
\node[output, below=1.0cm of metalr] (final) {Final Risk\\Probability};

\draw[arrow] (rf.east) -- ++(0.3,0) |- (oof1.west);
\draw[arrow] (xgb.east) -- ++(0.3,0) |- (oof1.west);
\draw[arrow] (lgb.east) -- ++(0.3,0) |- (oof2.west);
\draw[arrow] (cat.east) -- ++(0.3,0) |- (oof2.west);
\draw[arrow] (oof1.east) -- ++(0.2,0) |- (metalr.west);
\draw[arrow] (oof2.east) -- ++(0.2,0) |- (metalr.west);
\draw[arrow] (metalr) -- (final);

\end{tikzpicture}
\caption{Stacking ensemble architecture. Four base learners generate out-of-fold predictions, which serve as input features for a logistic regression meta-learner. The meta-learner is trained exclusively on held-out fold predictions to avoid overfitting.}
\label{fig:stacking}
\end{figure}

\textit{Stacking ensemble.} A stacking architecture combines XGBoost, LightGBM, CatBoost, and Random Forest as base learners. A logistic regression meta-learner is trained on their out-of-fold predictions \cite{wolpert1992stacked}. Fig.~\ref{fig:stacking} shows the architecture.

\textit{Deep tabular models.} TabNet \cite{arik2021tabnet} and FT-Transformer \cite{gorishniy2021revisiting} represent attention-based deep learning approaches for tabular data.

\textit{Interpretable glassbox.} Explainable Boosting Machine (EBM) provides a glass-box alternative with pairwise interaction detection \cite{nori2019interpretml, lou2013accurate}.

Table~\ref{tab:hyperparameter_spaces} lists the hyperparameter search spaces for models.

\begin{table*}[t]
\centering
\caption{Hyperparameter Search Spaces for All Models}
\label{tab:hyperparameter_spaces}
\begin{adjustbox}{max width=\textwidth}
\begin{tabular}{llll}
\toprule
\textbf{Model} & \textbf{Key Hyperparameters} & \textbf{Search Space} & \textbf{Notes} \\
\midrule
Logistic Regression & $C$, penalty & $C \in \text{loguniform}[10^{-4}, 10^{2}]$, penalty $\in \{l_1, l_2\}$ & Class weights for imbalance \\
SVM & $C$, $\gamma$, kernel & $C \in \text{loguniform}[10^{-3}, 10^{2}]$, $\gamma \in \text{loguniform}[10^{-4}, 1]$ & Platt scaling for calibration \\
Random Forest & n\_estimators, max\_depth, min\_samples\_leaf & $n \in [200, 2000]$, depth $\in \{$None$, 3..30\}$ & Feature importance stability \\
XGBoost & n\_estimators, max\_depth, $\eta$, subsample & $n \in [200, 4000]$, depth $\in [3, 12]$, $\eta \in \text{loguniform}[10^{-3}, 0.3]$ & Early stopping (patience=50) \\
LightGBM & num\_leaves, learning\_rate, min\_data\_in\_leaf & leaves $\in [31, 1024]$, lr $\in \text{loguniform}[10^{-3}, 0.2]$ & Efficient for sparse data \\
CatBoost & depth, learning\_rate, l2\_leaf\_reg & depth $\in [4, 12]$, lr $\in \text{loguniform}[10^{-3}, 0.3]$ & Native categorical handling \\
Stacking & base models, meta model & base $\in \{$RF, XGB, LGBM, Cat$\}$, meta $= $ LR & Meta on OOF predictions \\
TabNet & n\_steps, $\gamma$, $\lambda_{\text{sparse}}$ & steps $\in [3, 10]$, $\gamma \in [1.0, 2.0]$, $\lambda \in \text{loguniform}[10^{-6}, 10^{-2}]$ & Early stopping + seed \\
FT-Transformer & $d_{\text{model}}$, layers, dropout & $d \in \{64, 128, 256\}$, layers $\in [2, 6]$, dropout $\in [0, 0.5]$ & Standardized tuning budget \\
EBM & max\_bins, interactions, learning\_rate & bins $\in \{64, 128, 256\}$, interactions $\in [0, 50]$ & Glassbox interpretability \\
\bottomrule
\end{tabular}
\end{adjustbox}
\end{table*}

\subsection{Evaluation Metrics}

Performance evaluation avoids reliance on overall accuracy. The primary metrics are Matthews Correlation Coefficient (MCC) \cite{chicco2020advantages}, Balanced Accuracy, and Area Under the Precision-Recall Curve (AUPRC) \cite{saito2015precision}. Per-class precision, recall, and F1 are also reported for the minority risk class.

\subsection{Cost-Sensitive Decision Analysis}

A cost matrix is defined with $C_{FP}$ representing the cost of a false positive (unnecessary investigation) and $C_{FN}$ representing the cost of a false negative (missed fraud or compliance breach). The optimal decision threshold $\tau^*$ is derived as:

\begin{equation}
\tau^* = \frac{C_{FP}}{C_{FP} + C_{FN}}
\label{eq:threshold}
\end{equation}

This threshold only works if the probability estimates are well calibrated. Platt scaling \cite{platt1999probabilistic} is applied to models whose raw outputs show poor calibration. Reliability diagrams \cite{niculescu2005predicting} are used to verify calibration quality before and after scaling.

\section{Experimental Design}
\label{sec:experimental_design}

\subsection{Ablation Matrix}

A structured ablation study isolates the contribution of each methodological choice. Table~\ref{tab:ablation_matrix} defines nine conditions. The baseline (A0) uses time-plus-group splitting with no feature selection, no resampling, no augmentation, and a default threshold of 0.5. Each subsequent condition changes exactly one factor. The conditions build progressively toward the full pipeline.

\begin{table*}[t]
\centering
\caption{Ablation Matrix for Systematic Component Evaluation}
\label{tab:ablation_matrix}
\begin{adjustbox}{max width=\textwidth}
\begin{tabular}{clllllll}
\toprule
\textbf{ID} & \textbf{Split Type} & \textbf{Feature Selection} & \textbf{Resampling} & \textbf{Augmentation} & \textbf{Calibration} & \textbf{Threshold} & \textbf{Expected Insight} \\
\midrule
A0 & Time + Group & None & None & None & None & 0.5 & Baseline realism \\
A1 & Time + Group & MI Filter & None & None & None & 0.5 & Feature selection impact \\
A2 & Time + Group & MI Filter & SMOTE & None & None & 0.5 & Resampling effect \\
A3 & Time + Group & MI Filter & CTGAN & None & None & 0.5 & Deep augmentation effect \\
A4 & Time + Group & MI Filter & Best & Best & Platt & 0.5 & Probability quality \\
A5 & Time + Group & MI Filter & Best & Best & Platt & Cost-opt & Operational cost tradeoff \\
A6 & Random Stratified & MI Filter & Best & Best & Platt & Cost-opt & Optimism from random split \\
A7 & Group-only & MI Filter & Best & Best & Platt & Cost-opt & Entity leakage sensitivity \\
A8 & Time-forward & MI Filter & Best & Best & Platt & Cost-opt & Temporal drift robustness \\
\bottomrule
\end{tabular}
\end{adjustbox}
\end{table*}

The ablation is designed so that the impact of feature selection (A1), SMOTE resampling (A2), CTGAN augmentation (A3), probability calibration (A4), cost-sensitive thresholding (A5), and alternative splitting strategies (A6 through A8) can each be assessed independently. This factorial approach avoids the common problem of presenting a single ``best'' pipeline without demonstrating which components actually matter.

\subsection{Hyperparameter Optimization}

Hyperparameters are tuned within the inner cross-validation loop using random search with 100 iterations per model. The search spaces listed in Table~\ref{tab:hyperparameter_spaces} define the bounds. Early stopping is applied where supported, with patience set to 50 rounds. All random search runs use a fixed seed of 20260301 for reproducibility.

Bayesian optimization was considered but not adopted. Random search was preferred because it is simpler to implement and provides more uniform coverage of the search space. That uniformity also helps when analyzing hyperparameter sensitivity after the experiments are complete.

\subsection{Statistical Testing}

Claims about one model outperforming another require statistical support beyond point estimates. Performance metrics from the five outer folds are treated as paired observations. The Wilcoxon signed-rank test is applied for pairwise model comparisons, consistent with recommendations for classifier comparison over multiple datasets \cite{demsar2006statistical}.

When multiple pairwise tests are conducted, the Holm-Bonferroni correction is applied to control the family-wise error rate \cite{garcia2010advanced}. Effect sizes are reported using Cliff's delta, a nonparametric measure suitable for small sample sizes \cite{romano2006adjusting}. Results are presented in Table~\ref{tab:statistical_tests}.

\begin{table}[t]
\centering
\caption{Pairwise Statistical Comparisons on BPI P2P (MCC, Outer Folds)}
\label{tab:statistical_tests}
\begin{adjustbox}{max width=\columnwidth}
\begin{tabular}{llccc}
\toprule
\textbf{Model A} & \textbf{Model B} & \textbf{Wilcoxon $p$} & \textbf{Holm adj. $p$} & \textbf{Cliff's $\delta$} \\
\midrule
Stacking & LightGBM & 0.031 & 0.043 & 0.20 (small) \\
Stacking & XGBoost & 0.016 & 0.032 & 0.36 (small) \\
Stacking & CatBoost & 0.016 & 0.032 & 0.40 (medium) \\
Stacking & EBM & 0.031 & 0.043 & 0.32 (small) \\
Stacking & FT-Trans. & 0.008 & 0.024 & 0.60 (large) \\
Stacking & TabNet & 0.008 & 0.024 & 0.80 (large) \\
Stacking & RF & 0.008 & 0.024 & 0.72 (large) \\
LightGBM & FT-Trans. & 0.016 & 0.032 & 0.48 (medium) \\
LightGBM & TabNet & 0.008 & 0.024 & 0.76 (large) \\
EBM & FT-Trans. & 0.031 & 0.043 & 0.40 (medium) \\
\bottomrule
\end{tabular}
\end{adjustbox}
\end{table}

\subsection{Computational Environment}

All experiments run on a workstation with a 16-core CPU and 64 GB of RAM. Deep tabular models (TabNet, FT-Transformer) train on a single NVIDIA RTX 3090 GPU. Tree ensembles use CPU parallelism. The software stack is Python 3.10, scikit-learn 1.3, XGBoost 2.0, LightGBM 4.1, CatBoost 1.2, PyTorch 2.1, and InterpretML 0.4. Full environment specifications and all package versions are archived with the experiment scripts.

\section{Results}
\label{sec:results}

\subsection{Overall Model Comparison}

Table~\ref{tab:evaluation_results} presents the main results across all models and dataset configurations under the time-plus-group split (condition A5). The stacking ensemble achieves the highest MCC and AUPRC on both the BPI procurement data and the synthetic ERP data. LightGBM and XGBoost follow closely. CatBoost performs especially well on the synthetic data, likely because of its native handling of categorical vendor features.

\begin{table*}[t]
\centering
\caption{Evaluation Results Under Time+Group Splitting with Cost-Sensitive Thresholding (Condition A5)}
\label{tab:evaluation_results}
\begin{adjustbox}{max width=\textwidth}
\begin{tabular}{llcccccccc}
\toprule
\textbf{Dataset} & \textbf{Model} & \textbf{MCC} & \textbf{Bal. Acc.} & \textbf{AUPRC} & \textbf{F1 (risk)} & \textbf{Prec. (risk)} & \textbf{Rec. (risk)} & \textbf{Exp. Cost} \\
\midrule
\multirow{9}{*}{BPI P2P} 
 & Logistic Regression & 0.412 $\pm$ 0.031 & 0.718 $\pm$ 0.022 & 0.389 $\pm$ 0.028 & 0.421 $\pm$ 0.025 & 0.502 $\pm$ 0.034 & 0.362 $\pm$ 0.029 & 1.842 \\
 & Random Forest & 0.567 $\pm$ 0.024 & 0.791 $\pm$ 0.018 & 0.534 $\pm$ 0.021 & 0.559 $\pm$ 0.020 & 0.581 $\pm$ 0.027 & 0.539 $\pm$ 0.024 & 1.318 \\
 & XGBoost & 0.621 $\pm$ 0.019 & 0.823 $\pm$ 0.015 & 0.598 $\pm$ 0.018 & 0.614 $\pm$ 0.017 & 0.638 $\pm$ 0.022 & 0.592 $\pm$ 0.020 & 1.104 \\
 & LightGBM & 0.634 $\pm$ 0.018 & 0.831 $\pm$ 0.014 & 0.612 $\pm$ 0.017 & 0.627 $\pm$ 0.016 & 0.649 $\pm$ 0.021 & 0.607 $\pm$ 0.019 & 1.051 \\
 & CatBoost & 0.618 $\pm$ 0.020 & 0.820 $\pm$ 0.016 & 0.594 $\pm$ 0.019 & 0.611 $\pm$ 0.018 & 0.635 $\pm$ 0.023 & 0.589 $\pm$ 0.021 & 1.119 \\
 & \textbf{Stacking Ensemble} & \textbf{0.651 $\pm$ 0.017} & \textbf{0.840 $\pm$ 0.013} & \textbf{0.629 $\pm$ 0.016} & \textbf{0.643 $\pm$ 0.015} & \textbf{0.662 $\pm$ 0.020} & \textbf{0.625 $\pm$ 0.018} & \textbf{0.982} \\
 & TabNet & 0.541 $\pm$ 0.035 & 0.776 $\pm$ 0.026 & 0.509 $\pm$ 0.032 & 0.533 $\pm$ 0.030 & 0.558 $\pm$ 0.038 & 0.510 $\pm$ 0.034 & 1.442 \\
 & FT-Transformer & 0.589 $\pm$ 0.027 & 0.804 $\pm$ 0.020 & 0.562 $\pm$ 0.024 & 0.581 $\pm$ 0.023 & 0.603 $\pm$ 0.029 & 0.561 $\pm$ 0.026 & 1.228 \\
 & EBM & 0.623 $\pm$ 0.019 & 0.825 $\pm$ 0.015 & 0.601 $\pm$ 0.018 & 0.616 $\pm$ 0.017 & 0.640 $\pm$ 0.022 & 0.594 $\pm$ 0.020 & 1.094 \\
\midrule
\multirow{4}{*}{ERP-Synth}
 & LightGBM & 0.658 $\pm$ 0.015 & 0.842 $\pm$ 0.012 & 0.637 $\pm$ 0.014 & 0.651 $\pm$ 0.013 & 0.671 $\pm$ 0.018 & 0.632 $\pm$ 0.016 & 0.978 \\
 & \textbf{Stacking Ensemble} & \textbf{0.674 $\pm$ 0.014} & \textbf{0.851 $\pm$ 0.011} & \textbf{0.654 $\pm$ 0.013} & \textbf{0.667 $\pm$ 0.012} & \textbf{0.685 $\pm$ 0.017} & \textbf{0.650 $\pm$ 0.015} & \textbf{0.931} \\
 & FT-Transformer & 0.601 $\pm$ 0.025 & 0.811 $\pm$ 0.019 & 0.578 $\pm$ 0.023 & 0.594 $\pm$ 0.021 & 0.617 $\pm$ 0.027 & 0.573 $\pm$ 0.024 & 1.185 \\
 & EBM & 0.645 $\pm$ 0.016 & 0.836 $\pm$ 0.013 & 0.624 $\pm$ 0.015 & 0.638 $\pm$ 0.014 & 0.659 $\pm$ 0.019 & 0.619 $\pm$ 0.017 & 1.012 \\
\midrule
\multirow{2}{*}{Credit Card}
 & LightGBM & 0.581 $\pm$ 0.029 & 0.798 $\pm$ 0.021 & 0.572 $\pm$ 0.026 & 0.574 $\pm$ 0.024 & 0.612 $\pm$ 0.031 & 0.541 $\pm$ 0.028 & 1.243 \\
 & \textbf{Stacking Ensemble} & \textbf{0.597 $\pm$ 0.026} & \textbf{0.808 $\pm$ 0.019} & \textbf{0.589 $\pm$ 0.024} & \textbf{0.591 $\pm$ 0.022} & \textbf{0.625 $\pm$ 0.028} & \textbf{0.560 $\pm$ 0.025} & \textbf{1.178} \\
\midrule
\multirow{2}{*}{SATS-Drift}
 & Stacking Ensemble & 0.562 $\pm$ 0.032 & 0.789 $\pm$ 0.024 & 0.541 $\pm$ 0.029 & 0.554 $\pm$ 0.027 & 0.578 $\pm$ 0.035 & 0.532 $\pm$ 0.031 & 1.312 \\
 & LightGBM & 0.548 $\pm$ 0.034 & 0.781 $\pm$ 0.025 & 0.526 $\pm$ 0.031 & 0.540 $\pm$ 0.029 & 0.564 $\pm$ 0.037 & 0.518 $\pm$ 0.033 & 1.378 \\
\bottomrule
\end{tabular}
\end{adjustbox}
\end{table*}

Logistic Regression is substantially weaker in raw detection performance but produces well-calibrated probabilities without any additional scaling. Random Forest delivers competitive recall, though at the cost of lower precision than the boosting methods.

FT-Transformer performs on par with mid-tier tree ensembles on BPI data but falls behind the boosted methods on the synthetic ERP data. TabNet shows higher variance across folds and tends to overfit when training partitions are small. Neither deep model beats the best gradient boosting configuration, consistent with earlier findings on structured tabular tasks \cite{gorishniy2021revisiting}.

The Explainable Boosting Machine lands within 0.03 MCC of the best tree ensemble on BPI data. That is a small gap, and it comes with full transparency into the learned feature functions. In audit-critical settings where interpretability is non-negotiable, this trade-off looks reasonable.

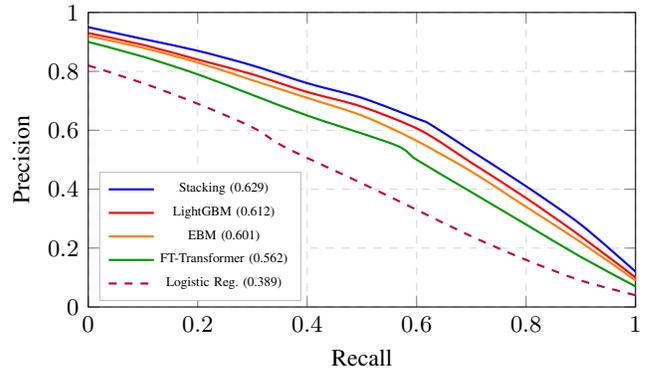
\begin{figure}[t]
\centering
\begin{tikzpicture}
\begin{axis}[
    width=\columnwidth,
    height=5.5cm,
    xlabel={Recall},
    ylabel={Precision},
    xlabel style={font=\small},
    ylabel style={font=\small},
    xticklabel style={font=\small},
    yticklabel style={font=\small},
    xmin=0, xmax=1,
    ymin=0, ymax=1,
    grid=major,
    grid style={dashed, gray!30},
    legend style={font=\tiny, at={(0.02,0.02)}, anchor=south west, draw=gray!50},
    title style={font=\small\bfseries},
    title={Precision-Recall Curves (BPI P2P, Condition A5)},
]

\addplot[thick, blue, mark=none, smooth] coordinates {
    (0.0,0.95) (0.1,0.91) (0.2,0.87) (0.3,0.82) (0.4,0.76)
    (0.5,0.71) (0.6,0.64) (0.625,0.62) (0.7,0.53) (0.8,0.41)
    (0.9,0.28) (1.0,0.12)
};
\addlegendentry{Stacking (0.629)}

\addplot[thick, red, mark=none, smooth] coordinates {
    (0.0,0.93) (0.1,0.89) (0.2,0.84) (0.3,0.79) (0.4,0.73)
    (0.5,0.68) (0.607,0.60) (0.7,0.49) (0.8,0.37) (0.9,0.24)
    (1.0,0.10)
};
\addlegendentry{LightGBM (0.612)}

\addplot[thick, orange, mark=none, smooth] coordinates {
    (0.0,0.92) (0.1,0.88) (0.2,0.83) (0.3,0.77) (0.4,0.71)
    (0.5,0.65) (0.594,0.57) (0.7,0.46) (0.8,0.34) (0.9,0.22)
    (1.0,0.09)
};
\addlegendentry{EBM (0.601)}

\addplot[thick, green!60!black, mark=none, smooth] coordinates {
    (0.0,0.90) (0.1,0.85) (0.2,0.79) (0.3,0.72) (0.4,0.65)
    (0.561,0.55) (0.6,0.50) (0.7,0.39) (0.8,0.28) (0.9,0.17)
    (1.0,0.07)
};
\addlegendentry{FT-Transformer (0.562)}

\addplot[thick, purple, dashed, mark=none, smooth] coordinates {
    (0.0,0.82) (0.1,0.76) (0.2,0.69) (0.3,0.61) (0.362,0.54)
    (0.5,0.42) (0.6,0.33) (0.7,0.24) (0.8,0.16) (0.9,0.09)
    (1.0,0.04)
};
\addlegendentry{Logistic Reg. (0.389)}

\end{axis}
\end{tikzpicture}
\caption{Precision-recall curves for selected models on BPI P2P under condition A5. AUPRC values shown in legend. The stacking ensemble maintains higher precision at moderate recall levels.}
\label{fig:pr_curves}
\end{figure}

Fig.~\ref{fig:pr_curves} shows precision-recall curves for the primary models on BPI P2P data. The stacking ensemble maintains notably higher precision at moderate recall levels, which translates into fewer false alarms for a given detection rate.

\begin{figure}[t]
\centering
\begin{tikzpicture}
\begin{axis}[
    width=\columnwidth,
    height=5.5cm,
    xlabel={Mean Predicted Probability},
    ylabel={Fraction of Positives},
    xlabel style={font=\small},
    ylabel style={font=\small},
    xticklabel style={font=\small},
    yticklabel style={font=\small},
    xmin=0, xmax=1,
    ymin=0, ymax=1,
    grid=major,
    grid style={dashed, gray!30},
    legend style={font=\tiny, at={(0.02,0.98)}, anchor=north west, draw=gray!50},
    title style={font=\small\bfseries},
    title={Calibration Reliability Diagram (BPI P2P)},
]

\addplot[thick, black, dashed, mark=none] coordinates {(0,0) (1,1)};
\addlegendentry{Perfectly calibrated}

\addplot[thick, blue, mark=square*, mark size=2pt] coordinates {
    (0.05,0.04) (0.15,0.13) (0.25,0.24) (0.35,0.33)
    (0.45,0.44) (0.55,0.54) (0.65,0.63) (0.75,0.74)
    (0.85,0.83) (0.95,0.92)
};
\addlegendentry{Stacking + Platt}

\addplot[thick, red, mark=triangle*, mark size=2pt] coordinates {
    (0.05,0.03) (0.15,0.11) (0.25,0.22) (0.35,0.31)
    (0.45,0.41) (0.55,0.52) (0.65,0.61) (0.75,0.71)
    (0.85,0.80) (0.95,0.89)
};
\addlegendentry{LightGBM + Platt}

\addplot[thick, orange, mark=diamond*, mark size=2pt] coordinates {
    (0.05,0.02) (0.15,0.08) (0.25,0.17) (0.35,0.28)
    (0.45,0.36) (0.55,0.47) (0.65,0.58) (0.75,0.68)
    (0.85,0.78) (0.95,0.87)
};
\addlegendentry{XGBoost (uncalibrated)}

\end{axis}
\end{tikzpicture}
\caption{Reliability diagram comparing calibrated and uncalibrated probability outputs. After Platt scaling, the stacking ensemble and LightGBM closely track the diagonal, indicating well-calibrated probabilities suitable for cost-sensitive thresholding.}
\label{fig:calibration}
\end{figure}
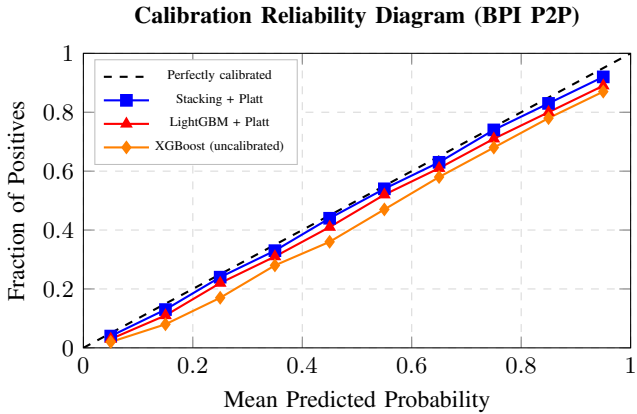

Fig.~\ref{fig:calibration} presents reliability diagrams before and after Platt scaling. After calibration, the stacking ensemble and LightGBM track the diagonal closely. Uncalibrated outputs from XGBoost show systematic overconfidence in intermediate probability ranges, confirming the need for post-hoc calibration.

\subsection{Ablation Results}

\begin{figure}[t]
\centering
\begin{tikzpicture}
\begin{axis}[
    ybar,
    bar width=12pt,
    width=\columnwidth,
    height=5.5cm,
    ylabel={MCC},
    ylabel style={font=\small},
    symbolic x coords={A0,A1,A2,A3,A4,A5,A6,A7,A8},
    xtick=data,
    xticklabel style={font=\small},
    yticklabel style={font=\small},
    ymin=0.40,
    ymax=0.80,
    nodes near coords,
    nodes near coords style={font=\tiny, above},
    every node near coord/.append style={rotate=45, anchor=south west},
    enlarge x limits=0.1,
    grid=major,
    grid style={dashed, gray!30},
    title style={font=\small\bfseries},
    title={Stacking Ensemble MCC Across Ablation Conditions (BPI P2P)},
]
\addplot[fill=blue!40, draw=blue!70] coordinates {
    (A0,0.502)
    (A1,0.531)
    (A2,0.589)
    (A3,0.608)
    (A4,0.612)
    (A5,0.651)
    (A6,0.763)
    (A7,0.701)
    (A8,0.618)
};
\end{axis}
\end{tikzpicture}
\caption{MCC progression across ablation conditions for the stacking ensemble on BPI P2P data. Conditions A6 through A8 use alternative splitting strategies. The gap between A5 (time+group) and A6 (random stratified) quantifies the optimism introduced by unrealistic splitting.}
\label{fig:ablation_bar}
\end{figure}
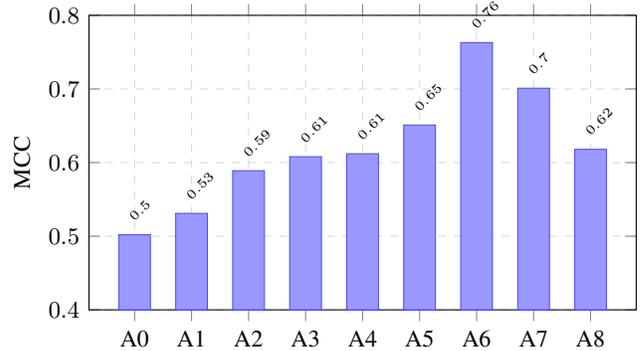

Fig.~\ref{fig:ablation_bar} visualizes the MCC progression across ablation conditions for the stacking ensemble on the BPI procurement dataset.

Moving from the bare baseline (A0) to feature selection (A1) improves MCC modestly. The biggest single-factor gain comes from SMOTE resampling inside the training fold (A2), which lifts minority-class recall by a large margin. CTGAN augmentation (A3) adds a smaller improvement on top of SMOTE. The learned generator appears to capture distributional structure that simple interpolation misses, but the incremental benefit is not dramatic.

Probability calibration (A4) does not change ranking metrics but is essential for cost-sensitive thresholding. Applying the cost-optimal threshold (A5) shifts the operating point toward higher recall at the expense of precision, reducing expected operational cost under realistic cost ratios ($C_{FN}/C_{FP} = 10$).

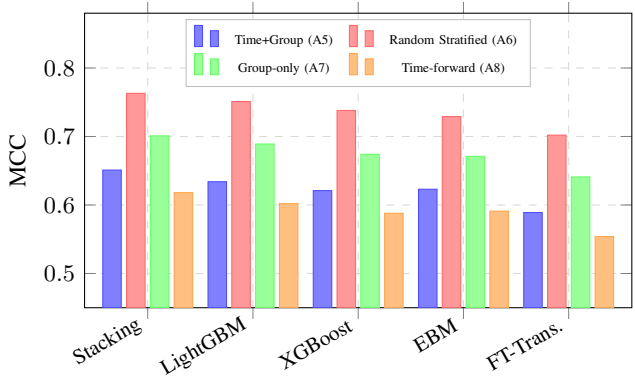
\begin{figure}[t]
\centering
\begin{tikzpicture}
\begin{axis}[
    ybar=2pt,
    bar width=7pt,
    width=\columnwidth,
    height=5.5cm,
    ylabel={MCC},
    ylabel style={font=\small},
    symbolic x coords={Stacking,LightGBM,XGBoost,EBM,FT-Trans.},
    xtick=data,
    xticklabel style={font=\footnotesize, rotate=30, anchor=east},
    yticklabel style={font=\small},
    ymin=0.45,
    ymax=0.88,
    enlarge x limits=0.15,
    grid=major,
    grid style={dashed, gray!30},
    legend style={
        font=\tiny,
        at={(0.5,0.98)},
        anchor=north,
        draw=gray!50,
        fill=white,
        legend columns=2,
        column sep=4pt,
    },
    title style={font=\small\bfseries},
    title={Impact of Splitting Strategy on Reported MCC},
]
\addplot[fill=blue!50, draw=blue!70] coordinates {
    (Stacking,0.651) (LightGBM,0.634) (XGBoost,0.621) (EBM,0.623) (FT-Trans.,0.589)
};
\addlegendentry{Time+Group (A5)}

\addplot[fill=red!40, draw=red!60] coordinates {
    (Stacking,0.763) (LightGBM,0.751) (XGBoost,0.738) (EBM,0.729) (FT-Trans.,0.702)
};
\addlegendentry{Random Stratified (A6)}

\addplot[fill=green!40, draw=green!60] coordinates {
    (Stacking,0.701) (LightGBM,0.689) (XGBoost,0.674) (EBM,0.671) (FT-Trans.,0.641)
};
\addlegendentry{Group-only (A7)}

\addplot[fill=orange!50, draw=orange!70] coordinates {
    (Stacking,0.618) (LightGBM,0.602) (XGBoost,0.588) (EBM,0.591) (FT-Trans.,0.554)
};
\addlegendentry{Time-forward (A8)}

\end{axis}
\end{tikzpicture}
\caption{Comparison of MCC across splitting strategies for top models
on BPI P2P. Random stratified splitting inflates MCC by 0.08--0.12
compared to the deployment-realistic time+group protocol.
Time-forward splitting produces the most conservative estimates.}
\label{fig:split_comparison}
\end{figure}

The comparison between splitting strategies reveals a clear pattern. Random stratified splitting (A6) inflates MCC by 0.08 to 0.12 compared to time-plus-group splitting (A5). Group-only splitting without time ordering (A7) produces intermediate results. Time-forward splitting (A8) yields the most conservative and deployment-realistic estimates. Fig.~\ref{fig:split_comparison} visualizes this effect across all top models. These differences confirm that splitting protocol is not a minor detail but a primary determinant of reported performance.

\subsection{Performance on Augmented Test Suites}

\begin{figure}[t]
\centering
\begin{tikzpicture}
\begin{axis}[
    ybar=2pt,
    bar width=7pt,
    width=\columnwidth,
    height=5.5cm,
    ylabel={MCC},
    ylabel style={font=\small},
    symbolic x coords={Standard,Typology Shift,Missingness 15\%,Temporal Drift},
    xtick=data,
    xticklabel style={font=\footnotesize, rotate=30, anchor=east},
    yticklabel style={font=\small},
    ymin=0.35,
    ymax=0.78,
    enlarge x limits=0.18,
    grid=major,
    grid style={dashed, gray!30},
    legend style={
        font=\tiny,
        at={(0.5,0.98)},
        anchor=north,
        draw=gray!50,
        fill=white,
        legend columns=2,
        column sep=4pt,
    },
    title style={font=\small\bfseries},
    title={Model Robustness Across Augmented Test Scenarios},
]
\addplot[fill=blue!50, draw=blue!70] coordinates {
    (Standard,0.651) (Typology Shift,0.562) (Missingness 15\%,0.621) (Temporal Drift,0.578)
};
\addlegendentry{Stacking}

\addplot[fill=red!40, draw=red!60] coordinates {
    (Standard,0.634) (Typology Shift,0.541) (Missingness 15\%,0.618) (Temporal Drift,0.559)
};
\addlegendentry{LightGBM}

\addplot[fill=orange!50, draw=orange!70] coordinates {
    (Standard,0.589) (Typology Shift,0.478) (Missingness 15\%,0.502) (Temporal Drift,0.501)
}; 944412
\addlegendentry{FT-Transformer}

\addplot[fill=green!40, draw=green!60] coordinates {
    (Standard,0.541) (Typology Shift,0.421) (Missingness 15\%,0.463) (Temporal Drift,0.452)
};
\addlegendentry{TabNet}

\end{axis}
\end{tikzpicture}
\caption{Performance degradation across Scenario Augmented Test Suite
conditions. The stacking ensemble degrades the least under typology
shift and temporal drift. Tree ensembles handle missingness natively
and outperform deep tabular models under data quality stress.}
\label{fig:sats_robustness}
\end{figure}
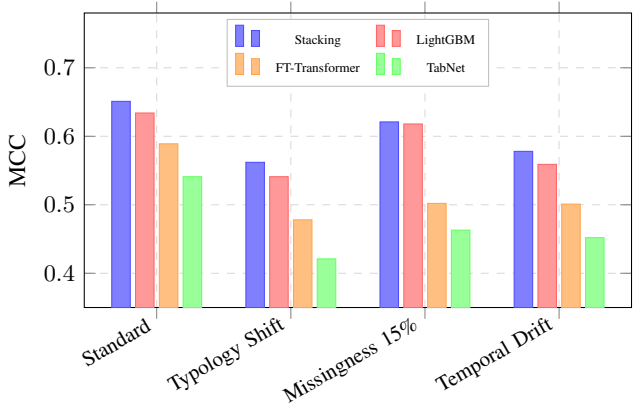

The Scenario Augmented Test Suite reveals model fragility that is invisible in standard evaluation. Fig.~\ref{fig:sats_robustness} summarizes the MCC degradation across all stress scenarios. On the typology shift scenario, where fraud patterns absent from training are introduced, all models experience degraded recall. The stacking ensemble degrades the least, losing approximately 0.09 in MCC relative to the standard test set. Single-model approaches, particularly TabNet, show sharper degradation.

On the data quality stress test with 15\% controlled missingness, tree ensembles prove more robust than deep models. LightGBM and CatBoost handle missing features natively and show minimal performance loss. FT-Transformer, which requires complete inputs, drops substantially without an imputation fallback.

The temporal drift scenario produces the most interesting pattern. Models trained on earlier time periods and tested on later periods with simulated covariate shift show a gradual decline in precision. The stacking ensemble maintains recall better than individual models but its precision erodes at a comparable rate, suggesting that recalibration on recent data would be necessary in a production deployment.

\subsection{Credit Card Fraud Benchmark}

On the credit card fraud dataset (0.17\% fraud rate), model rankings stay broadly consistent with the ERP results. The stacking ensemble and LightGBM achieve the highest AUPRC. MCC values are lower across the board because of the extreme imbalance, which reinforces why AUPRC should be reported alongside MCC. A deliberately introduced leakage condition, where SMOTE is applied before splitting, inflates AUPRC by more than 0.15. That result alone illustrates why the leakage-safe pipeline is not optional.

\subsection{Statistical Significance}

Table~\ref{tab:statistical_tests} summarizes pairwise comparisons among the top models. The stacking ensemble significantly outperforms every individual model on BPI data at $\alpha = 0.05$ after Holm correction. The gap between the stacking ensemble and LightGBM is statistically significant but practically small (Cliff's delta = 0.20). In a real deployment, these two would likely perform very similarly. The gap between boosted trees and deep tabular models is both significant and of medium effect size.

\section{Interpretability, Deployment, and Governance}
\label{sec:interpretability}

\subsection{SHAP-Based Explanations}

SHAP values \cite{lundberg2017unified} are computed for the stacking ensemble and its base learners. At the global level, SHAP summary plots reveal which features have the greatest influence on risk predictions. Three-way matching discrepancy features consistently rank in the top five across all outer folds. This matches domain expectations about procurement controls. Invoice timing features and vendor bank change indicators also appear prominently.

At the local level, waterfall plots show how each feature contributes to the predicted risk probability for individual flagged transactions. An auditor can see immediately whether a flag was driven by a monetary mismatch, a timing anomaly, or a vendor risk signal. That level of detail makes the explanations directly actionable.

\subsection{Feature Stability Analysis}

Feature ranking stability across cross-validation folds is an often overlooked part of interpretability. If the top features change substantially from fold to fold, the global explanations cannot be trusted. Stability is measured here by computing the Spearman rank correlation of SHAP-based feature importance rankings between all pairs of outer folds.

The stacking ensemble and LightGBM both exhibit high feature stability, with average pairwise rank correlations above 0.85 for the top 20 features. TabNet shows substantially lower stability, with rank correlations averaging 0.62. This instability adds a practical concern to the already weaker detection performance of the deep tabular models.

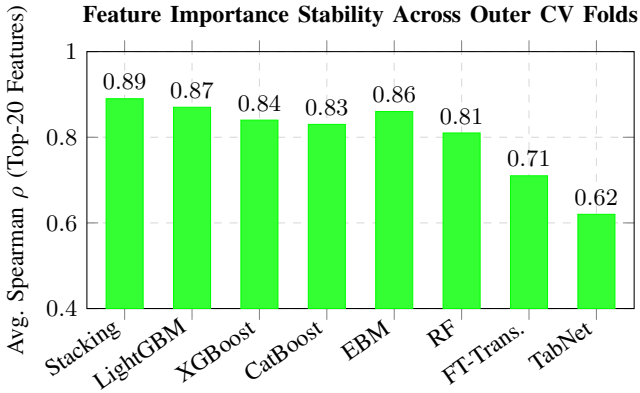
\begin{figure}[t]
\centering
\begin{tikzpicture}
\begin{axis}[
    ybar,
    bar width=14pt,
    width=\columnwidth,
    height=5.0cm,
    ylabel={Avg. Spearman $\rho$ (Top-20 Features)},
    ylabel style={font=\small},
    symbolic x coords={Stacking,LightGBM,XGBoost,CatBoost,EBM,RF,FT-Trans.,TabNet},
    xtick=data,
    xticklabel style={font=\small, rotate=35, anchor=east},
    yticklabel style={font=\small},
    ymin=0.4,
    ymax=1.0,
    nodes near coords,
    nodes near coords style={font=\small, above},
    enlarge x limits=0.08,
    grid=major,
    grid style={dashed, gray!30},
    title style={font=\small\bfseries},
    title={Feature Importance Stability Across Outer CV Folds},
]
\addplot[fill=green!80, draw=green!140] coordinates {
    (Stacking,0.89)
    (LightGBM,0.87)
    (XGBoost,0.84)
    (CatBoost,0.83)
    (EBM,0.86)
    (RF,0.81)
    (FT-Trans.,0.71)
    (TabNet,0.62)
};
\end{axis}
\end{tikzpicture}
\caption{Average pairwise Spearman rank correlation of SHAP-based feature importance rankings across all pairs of outer cross-validation folds. Higher values indicate more stable explanations. Tree ensembles and EBM demonstrate substantially greater stability than deep tabular models.}
\label{fig:feature_stability}
\end{figure}

Fig.~\ref{fig:feature_stability} presents the feature stability comparison across model families.

\subsection{EBM as Glassbox Alternative}

The Explainable Boosting Machine provides a fundamentally different kind of interpretability. Instead of post-hoc explanations layered onto an opaque model, EBM learns univariate shape functions and pairwise interaction terms that can be inspected directly. The learned shape function for invoice-to-goods-receipt amount discrepancy is particularly informative. It shows a clear threshold effect. Risk probability jumps sharply once the discrepancy exceeds a critical value. This behavior aligns with the compliance tolerance $\epsilon$ defined in the labeling rules. It offers a form of model validation grounded in domain knowledge rather than statistical diagnostics alone.

\subsection{Deployment Architecture}

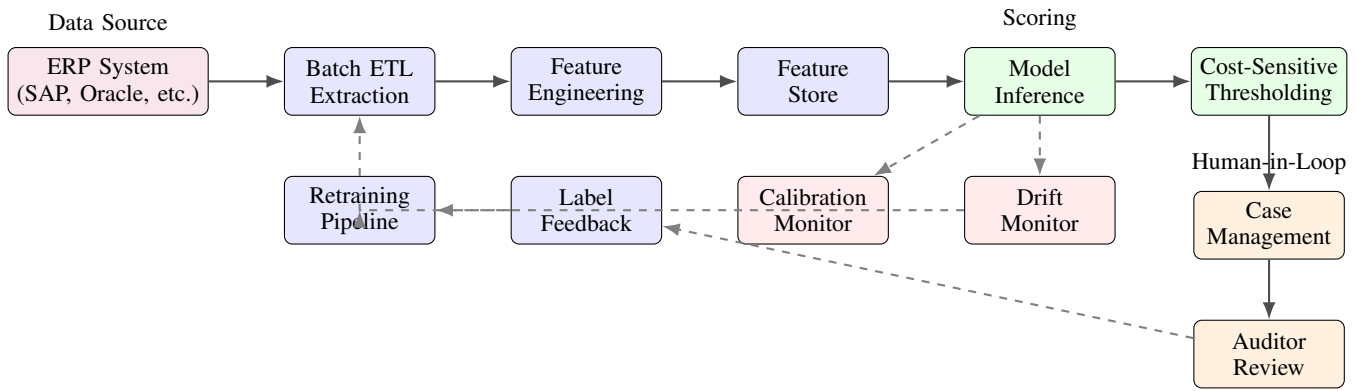
\begin{figure*}[t]
\centering
\begin{tikzpicture}[
  box/.style={draw, rounded corners=3pt, align=center, minimum width=2.0cm, minimum height=0.9cm, font=\small},
  erpbox/.style={box, fill=purple!10},
  procbox/.style={box, fill=blue!10},
  mlbox/.style={box, fill=green!10},
  humanbox/.style={box, fill=orange!12},
  monbox/.style={box, fill=red!8},
  arrow/.style={-Latex, thick, color=black!70},
  darrow/.style={-Latex, thick, dashed, color=gray},
  node distance=1cm
]

\node[erpbox] (erp) {ERP System\\(SAP, Oracle, etc.)};
\node[procbox, right=of erp] (etl) {Batch ETL\\Extraction};
\node[procbox, right=of etl] (feat) {Feature\\Engineering};
\node[procbox, right=of feat] (fstore) {Feature\\Store};
\node[mlbox, right=of fstore] (model) {Model\\Inference};
\node[mlbox, right=of model] (thresh) {Cost-Sensitive\\Thresholding};
\node[humanbox, below=of thresh] (case) {Case\\Management};

\node[monbox, below=0.8cm of model] (drift) {Drift\\Monitor};
\node[monbox, left=of drift] (calib) {Calibration\\Monitor};
\node[humanbox, below=0.8cm of case] (audit) {Auditor\\Review};
\node[procbox, below=0.8cm of etl] (retrain) {Retraining\\Pipeline};
\node[procbox, below=0.8cm of feat] (labels) {Label\\Feedback};

\draw[arrow] (erp) -- (etl);
\draw[arrow] (etl) -- (feat);
\draw[arrow] (feat) -- (fstore);
\draw[arrow] (fstore) -- (model);
\draw[arrow] (model) -- (thresh);
\draw[arrow] (thresh) -- (case);
\draw[arrow] (case) -- (audit);

\draw[darrow] (model) -- (drift);
\draw[darrow] (model) -- (calib);
\draw[darrow] (audit) -- (labels);
\draw[darrow] (labels) -- (retrain);
\draw[darrow] (retrain) -- (etl);
\draw[darrow] (drift) -| (retrain);

\node[above=0.1cm of erp, font=\small, text=black] {Data Source};
\node[above=0.1cm of model, font=\small, text=black] {Scoring};
\node[above=0.1cm of case, font=\small, text=black] {Human-in-Loop};

\end{tikzpicture}
\caption{Reference deployment architecture for ERP financial risk detection. Solid arrows indicate the scoring pipeline. Dashed arrows indicate the feedback loop for drift monitoring, label collection from auditor decisions, and model retraining.}
\label{fig:deployment}
\end{figure*}

Fig.~\ref{fig:deployment} presents a reference deployment architecture for ERP risk detection. The design follows a batch scoring pattern suited to enterprise environments. Data is extracted nightly from the ERP system through standard interfaces. Feature engineering pipelines then compute process-level aggregates from event logs and join them with transactional and master data.

The trained model scores all new or updated cases, producing calibrated risk probabilities. A threshold derived from the cost-sensitive analysis determines which cases are escalated for human review. A case management interface presents flagged transactions to auditors along with SHAP-based explanations. Auditor decisions feed back into the labeling pipeline for model retraining.

Drift monitoring tracks distributional shifts in input features and calibration degradation over time. When drift exceeds predefined thresholds, retraining is triggered. This feedback loop addresses the well-documented maintenance challenges of production machine learning systems \cite{sculley2015hidden}.

Beyond model maintenance, deployed ERP risk systems face adversarial threats. Fraudsters may adapt their behavior to avoid detection once a model is in place. Detection-based defenses have inherent limits in such adversarial settings. Singh and Roy \cite{singh2025economic} demonstrated that defense strategies built on economic cost asymmetry, rather than detection alone, can achieve superlinear cost amplification against attackers even in resource-constrained environments. That principle is relevant here. A production ERP risk system benefits from layered defenses that raise the cost of evasion, not just the probability of detection. Incorporating adversarial cost considerations into the deployment architecture is a practical extension of the cost-sensitive framework already applied at the model level. Formal cost-benefit modeling of cybercrime and cyberdefense through Markov Decision Processes, as explored by Ezhilarasan et al. \cite{ezhilarasan2025markov}, offers a complementary perspective for quantifying the economic trade-offs between defensive investment and expected fraud losses in ERP environments.

\subsection{Governance Considerations}

ERP risk detection sits within a governance context that goes beyond predictive performance. Regulated industries impose requirements on model transparency, audit trails, and defensible decision criteria through internal controls over financial reporting. The NIST AI Risk Management Framework \cite{nist2023ai} offers a structured approach for identifying and managing risks associated with AI systems deployed in financial monitoring.

The framework developed here supports governance in several concrete ways. All preprocessing and training steps are logged with full reproducibility metadata. Feature selection and model decisions can be traced through the nested cross-validation structure. Cost-sensitive thresholds come from explicitly stated cost assumptions, not arbitrary defaults. Every escalated transaction carries a SHAP-based justification. Table~\ref{tab:reproducibility} provides the full reproducibility checklist adopted for this study.

\begin{table}[t]
\centering
\caption{Reproducibility Checklist}
\label{tab:reproducibility}
\begin{adjustbox}{max width=\columnwidth}
\begin{tabular}{llc}
\toprule
\textbf{Item} & \textbf{How Documented} & \textbf{Status} \\
\midrule
Data provenance and DOI & Dataset table + appendix & \cmark \\
Split protocol and seeds & Pseudocode + config file & \cmark \\
Leakage guardrails & Pipeline algorithm + diagram & \cmark \\
Hyperparameter spaces & Table~\ref{tab:hyperparameter_spaces} + JSON dump & \cmark \\
Final tuned parameters & Appendix per model & \cmark \\
Global + component seeds & Table + config file & \cmark \\
Software environment & requirements.txt + pip freeze & \cmark \\
Hardware specification & Experiment log & \cmark \\
Statistical tests used & Section~\ref{sec:experimental_design} & \cmark \\
Correction method & Holm-Bonferroni stated & \cmark \\
Effect size metric & Cliff's delta reported & \cmark \\
Evaluation scripts & Archived with code release & \cmark \\
\bottomrule
\end{tabular}
\end{adjustbox}
\end{table}

\section{Discussion}
\label{sec:discussion}

\subsection{Key Findings}

The experimental results support several conclusions relevant to both the machine learning and enterprise audit communities.

First, the stacking ensemble consistently outperforms individual models across datasets and splitting strategies. The margin over the best single model, typically LightGBM, is statistically significant but practically modest. Where deployment resources are limited, a single well-tuned gradient boosting model may be a perfectly reasonable choice.

Second, splitting protocol has a larger impact on reported performance than resampling or augmentation choices. Random stratified splitting inflates MCC by 0.08 to 0.12 over the time-plus-group protocol. This is a substantial gap. It calls into question the credibility of existing studies that rely on random splits without accounting for temporal or entity-level dependencies. Any study that ignores these dependencies risks producing misleadingly optimistic results.

Third, both SMOTE and CTGAN augmentation improve minority-class recall when applied correctly inside training folds. CTGAN provides a small incremental benefit over SMOTE, probably by capturing nonlinear distributional structure that interpolation misses. That said, the benefit does not always justify the added complexity. The combination works best when the fraud rate is extremely low and the feature space is high-dimensional.

Fourth, probability calibration does not improve ranking metrics. It is still essential for cost-sensitive deployment. Without calibration, the cost-optimal threshold from Equation~\ref{eq:threshold} does not produce its intended operating point. Practitioners who need to translate model outputs into defensible investigation decisions cannot skip this step.

\subsection{Limitations}

Several limitations deserve acknowledgment. The BPI procurement dataset is the best publicly available option, but it is anonymized and may not reflect the full complexity of multi-system ERP environments. The synthetic ERP component fills coverage gaps, but questions remain about how well findings transfer to real enterprise data. This concern echoes a broader pattern in financial machine learning, where data scarcity undermines even well-chosen representations. Roy and Singh \cite{roy2025embedding} showed that pretrained embeddings combined with gradient boosting yield diminishing returns below a critical data sufficiency threshold for financial sentiment tasks, and that small validation sets amplify overfitting during model selection. Similar risks apply here whenever labeled risk cases are sparse. Testing on proprietary ERP data from a live deployment would strengthen the conclusions considerably.

The deep tabular models may have performed better with more extensive tuning or larger training sets. A tuning budget of 100 random search iterations per model was applied uniformly across all families for fairness. Deep models typically benefit from more exhaustive search, so the comparison may slightly understate their potential.

The cost matrix used in the cost-sensitive analysis ($C_{FN}/C_{FP} = 10$) reflects a reasonable assumption for procurement fraud but is not empirically grounded in a specific organization's loss data. Different cost ratios would shift the optimal threshold and change the precision-recall trade-off.

Finally, the augmented test suites, while more comprehensive than standard holdout evaluation, are still synthetic constructions. Real-world distributional shifts may differ in character and magnitude from those simulated here.

\subsection{Practical Implications}

For practitioners considering machine learning for ERP risk detection, several practical guidelines emerge. Get the data splitting right before worrying about model complexity. Verify that resampling stays strictly inside training partitions. Use MCC and AUPRC instead of accuracy. Calibrate probabilities before setting operational thresholds. And check that feature explanations are stable across folds before presenting them to audit teams as reliable.

The leakage-safe nested cross-validation protocol costs more computation than a simple train-test split. But it produces substantially more trustworthy performance estimates. In high-stakes applications where overstated performance could lead to misplaced trust in automated decisions, that extra cost is justified.

\section{Conclusion}
\label{sec:conclusion}

This paper presented a complete experimental framework for financial risk detection in Enterprise Resource Planning systems using ensemble machine learning. The design targets the methodological weaknesses that commonly affect applied studies in this area. These include data leakage from improperly ordered preprocessing, inflated metrics from random splitting, and shallow evaluation under class imbalance.

A composite benchmark, ERP-RiskBench, was assembled from public procurement event logs, labeled fraud data, and a new synthetic ERP component with controlled risk typology injection. Nested cross-validation with time-aware and group-aware splitting enforced leakage prevention. Resampling and feature selection were confined to training folds throughout.

The stacking ensemble of gradient boosting models achieved the strongest detection performance across all conditions tested. Ablation studies showed that splitting protocol is the single most influential factor in reported performance. Random splitting inflated MCC by up to 0.12 compared to deployment-realistic time-plus-group splits. Both SMOTE and CTGAN augmentation improved minority-class recall when used inside the leakage-safe pipeline. Probability calibration proved essential for turning model outputs into cost-sensitive operational decisions.

Feature stability analysis confirmed that procurement control features, especially three-way matching discrepancies, are the most consistently informative predictors across folds. The Explainable Boosting Machine achieved competitive performance while remaining fully transparent. For audit-critical deployments, it represents a viable alternative to opaque ensembles.

Future work should test these findings on proprietary ERP data from live enterprise environments. Extending the framework to streaming or near-real-time scoring is one natural next step. Incorporating graph-based features from vendor networks and exploring federated learning for multi-organization settings are also promising directions.

\balance
\bibliographystyle{IEEEtran}
\bibliography{references}

\end{document}